\documentclass[conference]{IEEEtran}
\usepackage{times}

\usepackage{color}
\usepackage{graphicx}
\usepackage[bookmarks=true]{hyperref}
\usepackage{multicol}
\usepackage[numbers]{natbib}
\usepackage{subcaption}
\usepackage[usenames,dvipsnames]{xcolor}
\usepackage{verbatim}

\definecolor{YaleBlue}{HTML}{00268d}

\definecolor{green}{RGB}{3,112,15}

\usepackage{fancyhdr}

\begin{document}

\title{Incorporating Gaze into Social Navigation}

\author{Justin Hart, Reuth Mirsky, Xuesu Xiao, Peter Stone \\ Paper \#8}



%

\maketitle

\begin{abstract}
Most current approaches to social navigation focus on the trajectory and position of participants in the interaction. Our current work on the topic focuses on integrating gaze into social navigation, both to cue nearby pedestrians as to the intended trajectory of the robot and to enable the robot to read the intentions of nearby pedestrians. This paper documents a series of experiments in our laboratory investigating the role of gaze in social navigation.
\end{abstract}

\IEEEpeerreviewmaketitle

\section{Introduction}
As mobile robots move into human-populated environments, such as homes, offices, and businesses, they must be able to negotiate the problem of navigating in spaces that they share with people. This development has given rise to research on the problem of social navigation \cite{charalampous2017recent,kruse2013human}. Among other goals, researchers in this area wish to improve the comfort and safety of people who must share space with robots, to make robots more interpretable to people as they navigate, and to enable robots to make progress on tasking where they may otherwise be impeded by nearby pedestrians blocking their path \cite{bera2017sociosense, takayama2011expressing}. It is also worth noting that the study of social navigation has not been limited to the domain of robotics. Significant research has been performed in virtual reality simulations or 3D-rendered game models \cite{musse1997model,strassner2005virtual}, and the Social Force Model --- which has been leveraged in robotics --- has its origins in multi-agent crowd simulations \cite{helbing1995social, ratsamee2012modified}.

A significant majority of the work on the task of social navigation in robots has focused on the position and trajectory of people with respect to the robot \cite{chen2017socially, henry2010learning, okal2016learning, sisbot2007human}. A smaller collection of work has focused on cuing nearby pedestrians as to the intentions of the robot. Methods have included the addition of turn signals to the robot, as well as projection mapping arrows onto the floor in front of the robot; both indicating the robot's intended trajectory \cite{Baraka2018, fernandez2018passive, Ryo2021Omni, nummenmaa2009ll}. Research in our group has instead focused on leveraging gaze as a social cue, both to indicate the intended trajectory of the robot \cite{hart2020hallway} and to interpret the intentions of nearby people \cite{holman2021watch}. This short paper discusses the evolution of our thinking on this problem based on the outcomes of several experiments in an ongoing series of studies that we are performing and concludes with a discussion of challenges we have identified. For a comprehensive review of approaches for handling interactions in the context of social navigation, we refer the reader to Mirsky et al. \cite{mirsky2021prevention}.

The first work in this series of experiments is by Fernandez et al. \cite{fernandez2018passive} who attached LEDs to the frame of a BWIBot robot \cite{khandelwal2017bwibots}, and used the LEDs in a fashion similar to a turn signal. In a test in which people pass robots heading head-on towards them in a hallway, 
LEDs were only successful in preventing a human and a robot from blocking each other's paths when the person has previously seen the turn signal being used by the robot (thus revealing the signal's meaning). This result caused us to look to the use of gaze as a social cue to coordinate hallway-passing behavior, with the hypothesis that gaze will be easily interpreted correctly by people.

Gaze is an important indicator of where a person is about to move. Norman \cite{norman2009design} speculated that bicycle riders avoid collisions with pedestrians by reading their gaze. Nummenmaa et al. \cite{nummenmaa2009ll} present a study in which a virtual agent (3D-rendered on a computer monitor) walks towards the study participant. The participant must choose whether to pass the agent on its left or right using keyboard commands, and the virtual agent indicates its intention by looking to its left or right. Unhelkar et al. \cite{unhelkar2015human} present a study in which head pose is used to determine which target a pedestrian will walk toward. Khambhaitia et al.  \cite{khambhaita2016head} present a motion planner which coordinates the head motion of a robot to the path that the robot will take 4 seconds in the future, and asked participants in a video survey to determine the robot's intended path as it approaches a T-intersection.

In our lab we have investigated the use of gaze in social navigation. Hart et al. \cite{hart2020hallway} present a human study in which researchers acting as pedestrians in a busy hallway vary their gaze patterns to be either congruent with the direction that they intend to walk, counter to that direction, or absent (by looking down at a cell phone). The results demonstrate that pedestrians passing these researchers in a hallway are more likely to collide with them when their gaze is counter to the direction that they intend to walk in. In the same paper, Hart et al. \cite{hart2020hallway} update the experimental setup from Fernandez et al. \cite{fernandez2018passive} to compare a gaze cue presented on a 3D-rendered virtual agent head mounted to the robot's chassis to the use of LEDs; finding that the gaze cue is more effective than the LEDs in preventing the robot and participants in the study from blocking each other's paths.

The first two of these robot studies surrounds the idea of the robot socially cuing its intentions to nearby pedestrians, but does not explore the idea of the robot's behavior responding to social cues made by the people it interacts with \cite{gockley2007natural, tamura2010smooth}. The most recent study on this topic by our group, performed by Holman et al. \cite{holman2021watch}, approaches this problem from the perspective of enabling the robot to respond to human gaze. Participants are placed in a virtual environment, using wireless virtual reality equipment with an embedded eye tracker, and instructed to walk to one of five targets, in a similar fashion to the experimental design described in the work by Unhelkar et al. \cite{unhelkar2015human}. The study's findings indicate that gaze can be used as an early cue indicating the target of a participant's motion. We plan to leverage these results in future work to enable a robot to coordinate its motion to that of nearby pedestrians.


\section{Gaze, Navigation, and Hallway Passing}
In this section we provide further details on our experiments on gaze and social navigation.

\subsection{LEDs and Passive Demonstrations}
\begin{figure}
    \centering
    \includegraphics[width=4cm, height=6.5cm]{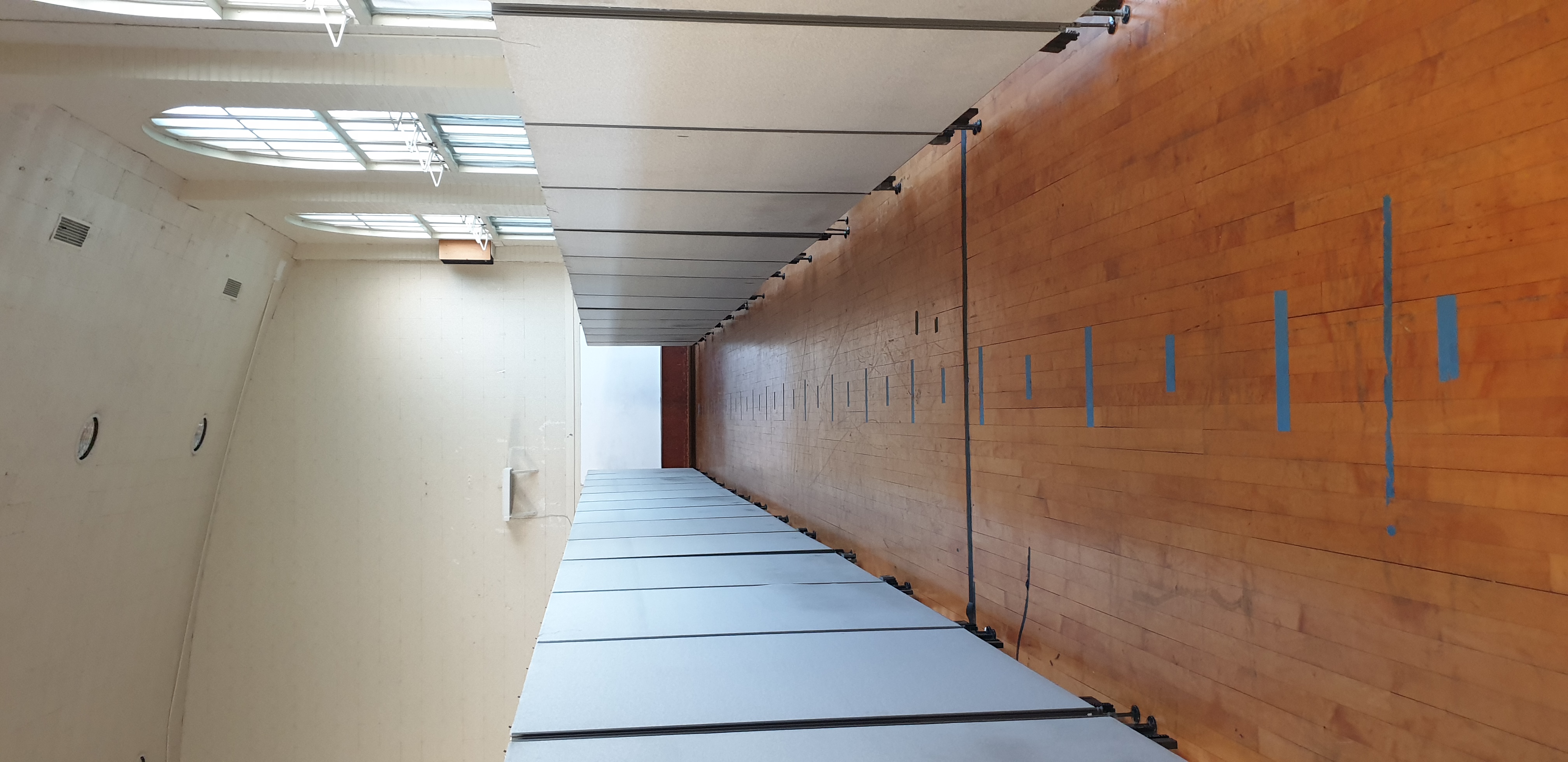}
    \includegraphics[width=4cm, height=6.5cm]{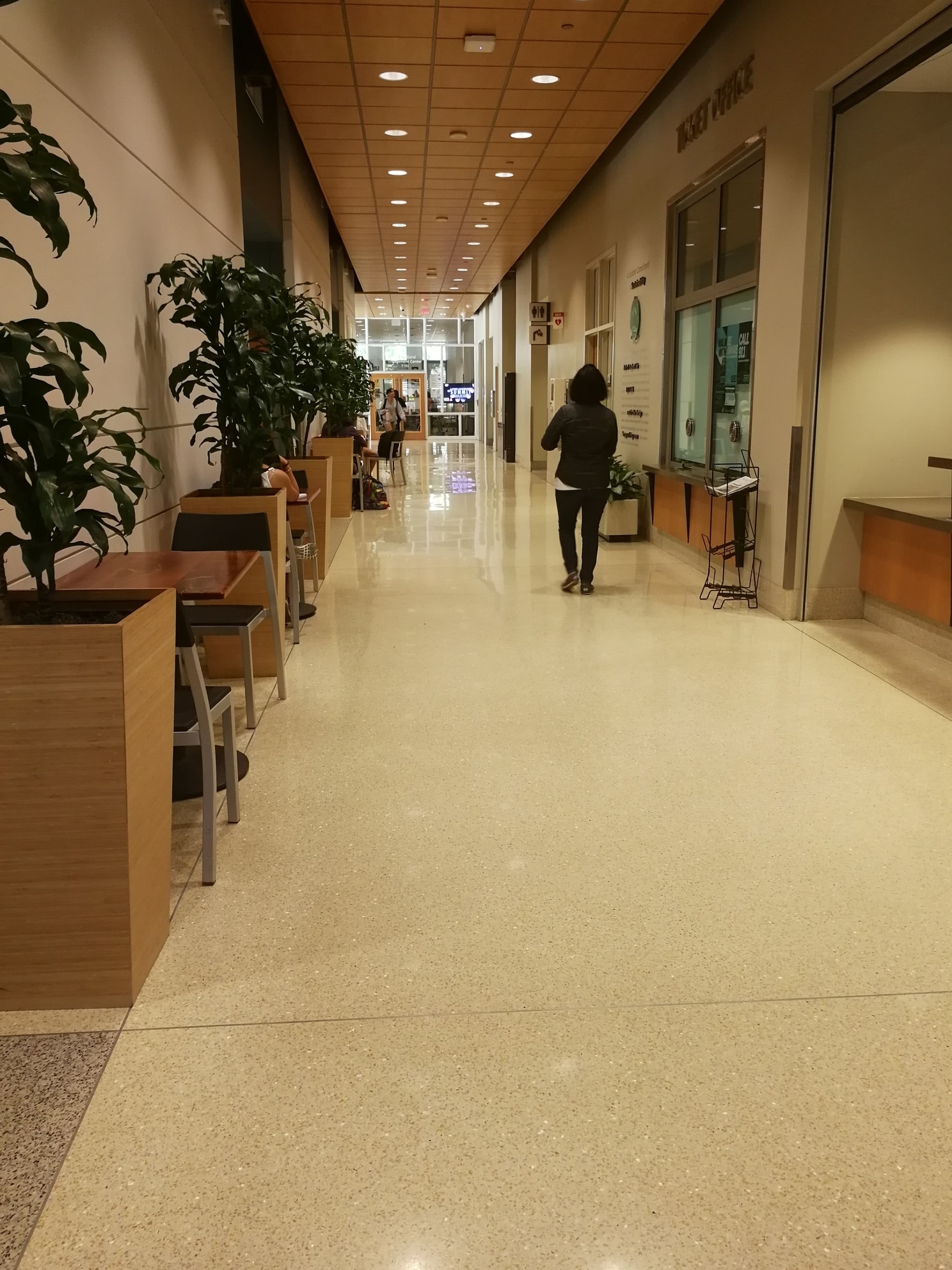}
    \caption{The hallway on the left is the one used in the human-robot interaction studies by Fernandez et al. \cite{fernandez2018passive} and Hart et al. \cite{hart2020hallway}. The one on the right is the one used in the human field study by Hart et al. \cite{hart2020hallway}.}
    \label{fig:hallways}
\end{figure}

\begin{figure}
    \centering
    \includegraphics[width=4cm]{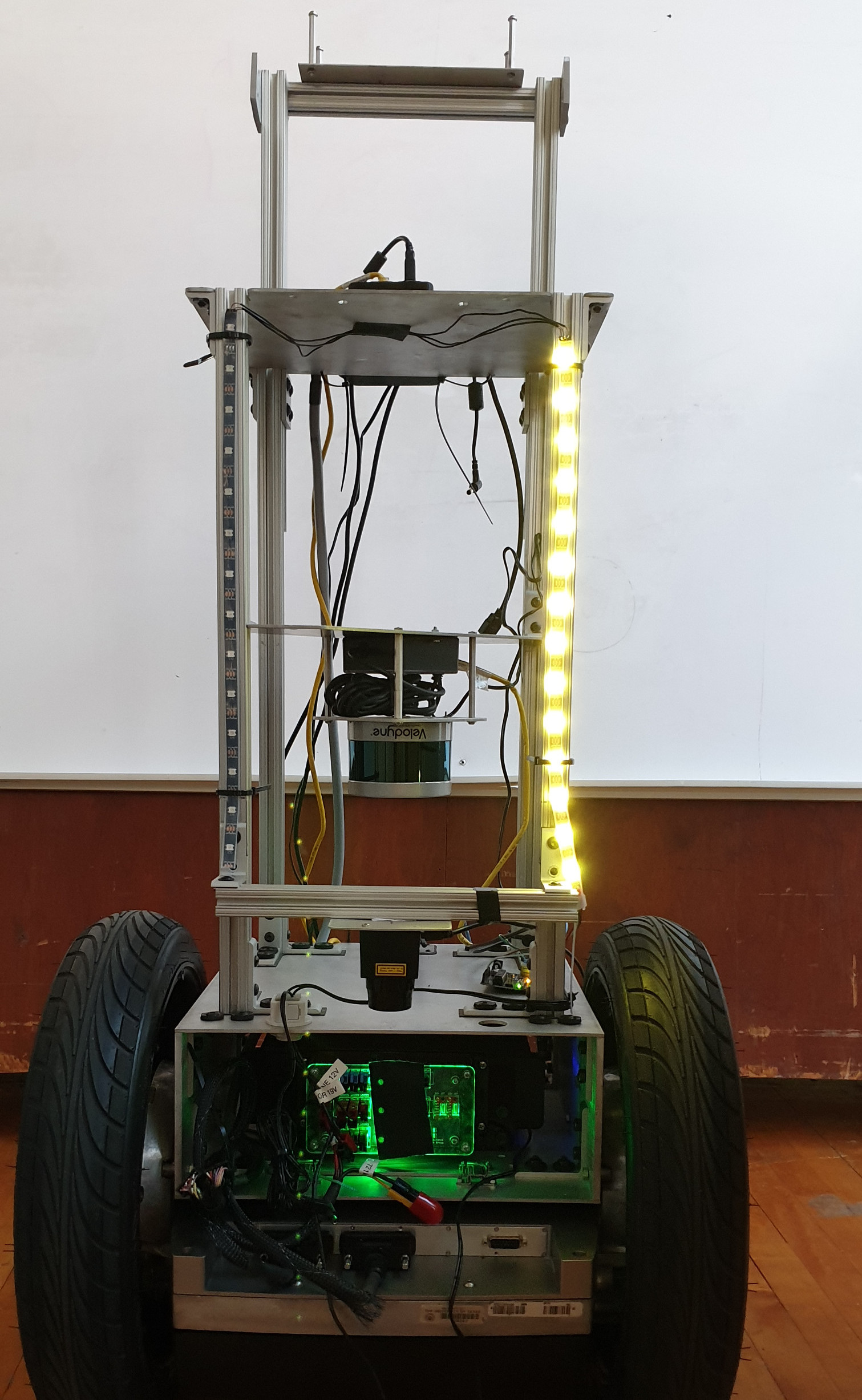}
    \includegraphics[width=4cm]{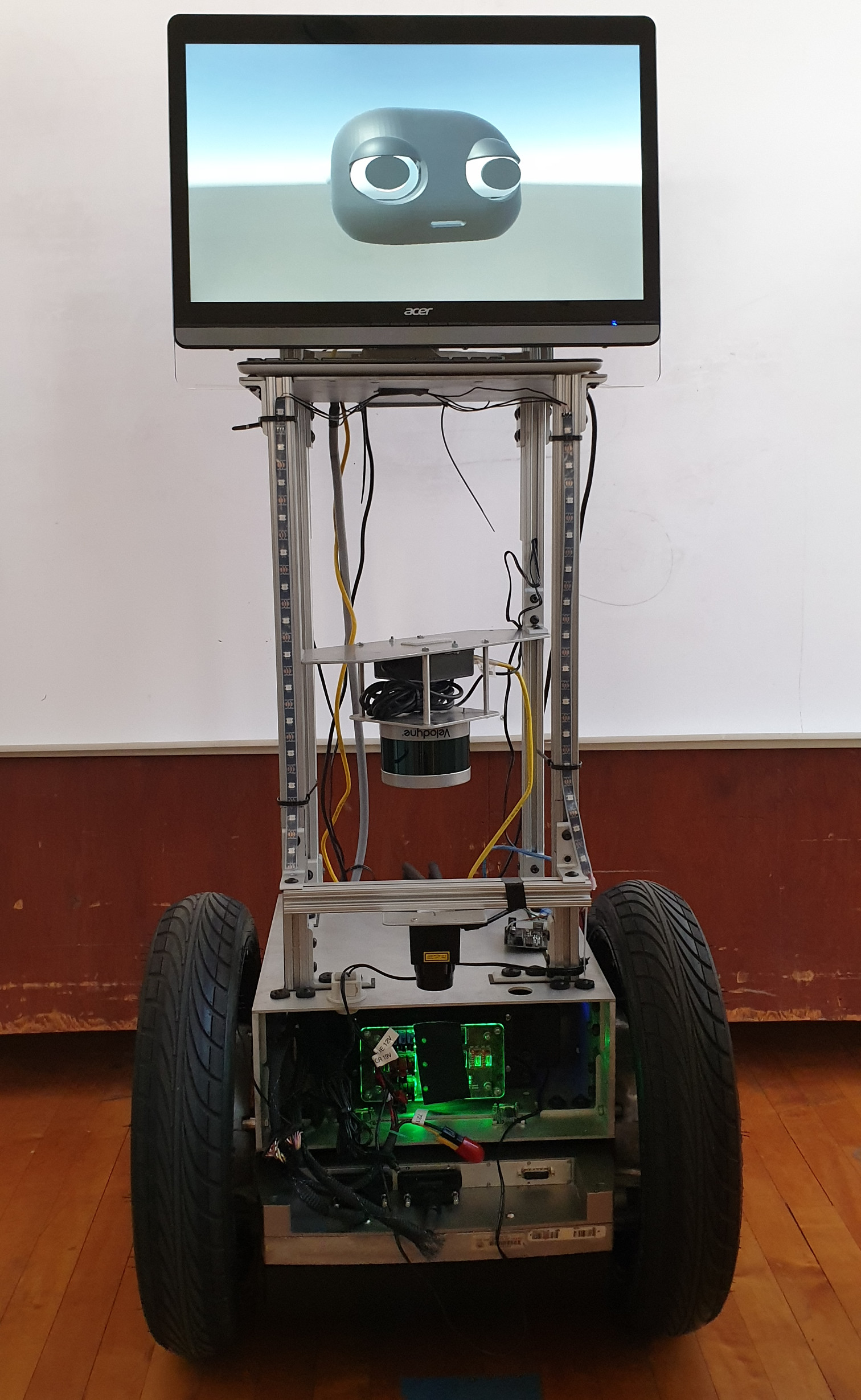}
    \caption{The two conditions in our human-robot hallway experiment: the LED signal (left) and the gaze signal (right).}
    \label{fig:exp}
\end{figure}

Fernandez et al \cite{fernandez2018passive} present a study in which a robot navigates a hallway (Figure~\ref{fig:hallways} (left)) and signals the side that it intends to pass a human participant using a strip of LEDs which act as a turn signal. The robot’s navigation algorithm treats the hallway as being divided into three traffic lanes through which it may navigate. Both the robot and the pedestrian start on the middle lane at opposite ends of a hallway. The robot signals that it is about to change its lane by blinking the LED light strip on the side of its chassis matching that of the direction of the lane that it intends to shift into. The LEDs are configured similarly to Figure \ref{fig:exp} (left), which is adapted  from \cite{hart2020hallway}. It should be noted that there is an important difference between the appearance of the robot with LEDs in \cite{fernandez2018passive} and \cite{hart2020hallway}. In Ferndandez et al. \cite{fernandez2018passive}, the robot has a monitor attached to its top in the LED condition, but with no face rendered on it, and mounted facing the back of the robot. This is a design choice on the BWIBot used to launch the robot's software. In the Hart et al. \cite{hart2020hallway} study, the monitor is removed because it was noted by the researchers that study participants would sometimes pause to observe the contents of the monitor, which is only the output of the ROS nodes driving the robot, and not intended as part of the interaction.

\begin{figure}
    \centering
    \includegraphics[width=0.45\textwidth]{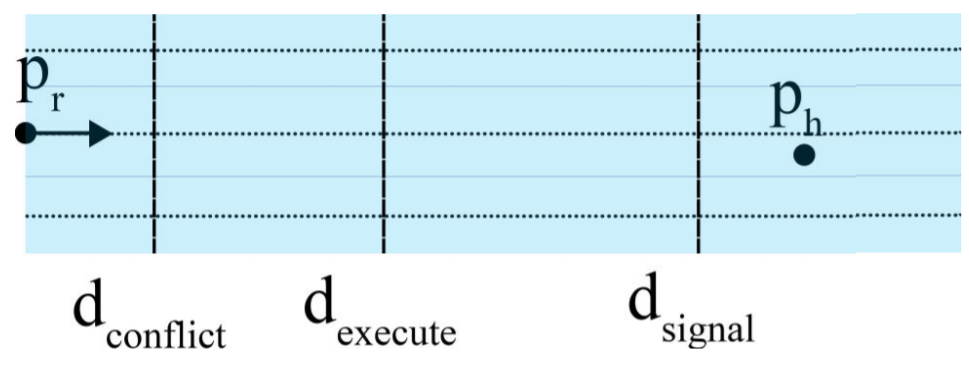}
    \caption{A diagram of the hallway, its lanes, and the distance thresholds at which the robot signals its intention to change lanes ($d_{conflict}$), executes a lane change ($d_{execute}$), and is determined to potentially be in conflict with a person in its path ($d_{signal}$). The position of the robot is marked $p_r$ and the position of the person is marked $p_h$.}
    \label{fig:lanes}
    \vspace{-10pt}
\end{figure}

The robot's navigation algorithm models passing a person in a hallway as a problem over three traffic lanes at three distances, as in Figure \ref{fig:lanes}. If the person and the robot are both in the middle lane, then the robot has the option of passing the person on the left or the right, by shifting into the corresponding lane. The distances in this model are: $d_{signal}$, the distance at which the robot will begin to signal its intention to the pedestrian; $d_{execute}$, the distance at which the robot will begin to shift into the left or right lane; and $d_{conflict}$, at which the robot stops its motion and does not attempt to pass. If the robot and the person are in opposite lanes when they pass each other, the robot will not stop during the interaction. 

In addition to these parameters, Fernandez et al. \cite{fernandez2018passive} introduced the concept of a ``passive demonstration,'' which is a sort of training episode in which the study participant is not informed that they are being trained, but wherein the robot demonstrates the signal by simply using it in front of the participant before it is relevant to their interaction. In this case, the robot moves into the right lane, using the turn signal, at the very start of crossing the hallway, then moves back to the middle prior to passing the participant. Upon coming within distance $d_{signal}$ the robot again signals, now moving into the left lane, when passing the participant.

The study follows an inter-participant design, in which each participant traverses the hallway exactly once. The distances, $d_{signal}$, $d_{execute}$, and $d_{conflict}$ are set to $7$ meters, $2.75$ meters, and $1$ meter, respectively, and the robot always moves into the left lane when passing the pedestrian. These values are chosen based on pilot study data, indicating that pedestrians are likely pass on the right, and that $d_{execute}$ is at the last possible distance change lanes. This set of distances is chosen to assure that participants who successfully pass the robot do so based on the signal, rather than the robot's motion. The study is set up as a $2X2$ experiment where the controlled variables are whether or not the LED is used, and whether or not the robot performs a passive demonstration. The main measure is whether a participant and the robot experience a ``conflict,'' in which they come too close to each other. The results show that the passive demonstration condition with the LED turn signal significantly outperforms other conditions (no demonstration, no LED: $100\%$ conflict; no demonstration, LED: $90\%$ conflict; demonstration, no LED: $70\%$ conflict; demonstration, LED: $20\%$ conflict). A one-way ANOVA shows a significant main effect ($F(3,36) = 9.913, p < 0.001)$) and all pairwise post-hoc tests based on Least Squares Difference (LSD) contrasting against the ``demonstration, LED'' condition are significant at $p < 0.01$.

\subsection{Gaze in Purely Human Navigation Environments}
The results from the previous experiment encouraged us to search for a more naturalistic cue that people would be able to pick up on without having to observe passive demonstrations. Following previous work that looked at potential cues such as body rotation, trajectory estimation, and gaze \cite{Patla1999, unhelkar2015human}, we conducted a human study where we tested the viability of gaze as an intentional cue in purely human navigation \cite{hart2020hallway}.

In this study, a researcher navigates the hallway depicted in Figure \ref{fig:hallways} (right) and looks either in the direction in which they intend to go (\textit{Congruent gaze}), opposite to this direction (\textit{Incongruent gaze}), or at a mobile phone to deprive other pedestrians from leveraging their gaze (\textit{No gaze}). The primary metric is whether the researcher comes into conflict with other pedestrians, defined as bumping into them, brushing against them, or quickly shifting to get out of each other's way.

A total of $220$ interactions were observed with $60$ congruent gaze interactions, $85$ incongurent gaze interactions, and $75$ no gaze interactions. The mean percentage of conflicts by condition are: congruent gaze, $15\%$; incongruent gaze, $48\%$; and no gaze $28\%$. A one-way ANOVA shows a significant main effect ($F_{2,217}=5.02, p=0.007)$. Post-hoc tests of pairwise mean differences using the Bonferroni criteria show significant differences between congruent gaze and the other two conditions (congruent vs. incongruent $md = 0.221, p=0.017$; congruent vs. no gaze $md=0.191, p=0.033$).

These results highlight the importance of gaze as a naturalistic cue that assists people to process the navigational goal of other pedestrians around them, and adapt their own trajectory accordingly. This outcome has motivated our subsequent studies on how gaze can be leveraged both to convey the robot's navigational goal and to infer the human navigational goal.

\subsection{Conveying the Robot's Navigational Intention}
Returning to the hallway used in the passive demonstration experiment \cite{fernandez2018passive}, we hypothesized that the use of gaze-like cue is more readily interpretable than the LED signal.

We designed a gaze cue using a 3D-rendered version of the Maki 3D-printable robot head.\footnote{https://www.hello-robo.com/} The virtual head is displayed on a $21.5$ inch monitor mounted to the front of the robot. When signaling, the robot turns its head $16.5^{\circ}$ and remains in this pose, as shown in Figure \ref{fig:exp} (right). The experimental design repeats the experimental setup from Fernandez et al. \cite{fernandez2018passive}, contrasting the gaze signal against the LED signal, but omitting the test of passive demonstrations. Because of hardware changes, $d_{signal}$ is reduced to $4$m.

With $11$ participants in the LED condition and $16$ in the gaze condition, participants in the gaze condition successfully infer the robot's goal $50\%$ of the time, while none of the participants in the LED condition infer the goal of the robot. Important to note in the interpretation of the results from this experiment is that we expect a conflict $100\%$ of the time unless the robot's cue (either LED or gaze) are correctly interpreted. This is because the robot moving into the left-hand lane, which is against the convention in North America, is expected to result in conflict 100\% of the time.
Comparing the performance of the gaze signal against the LED signal demonstrates its ability to impact people’s navigational choices, and that people more easily interpret the gaze cue than the LED turn signal.

\subsection{Inferring the Pedestrian's Navigational Intention}

 \begin{figure*}[htb!]
    \centering
    \includegraphics[width=0.9\linewidth]{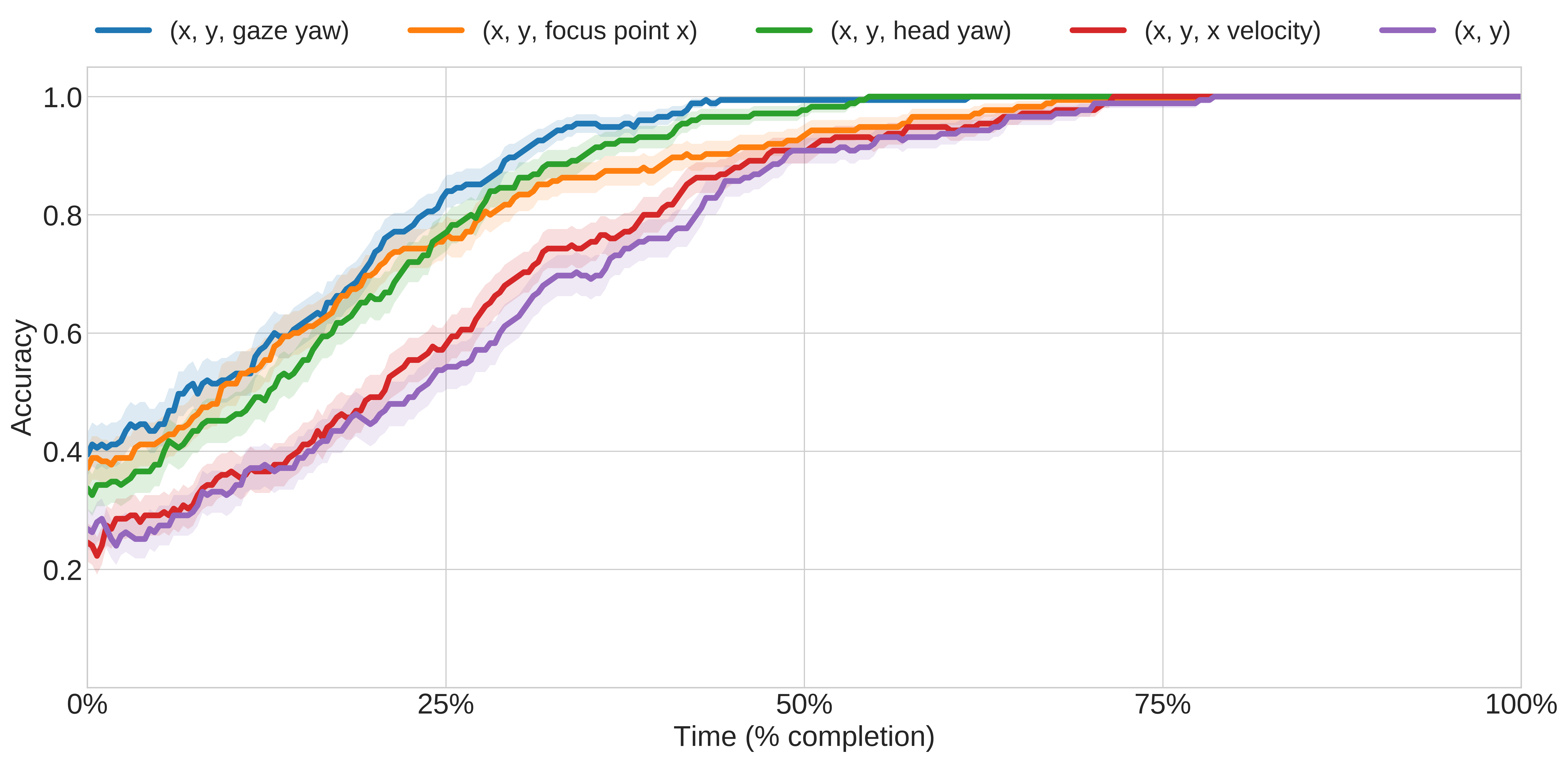}
    \caption{Cross-validated accuracy of the multivariate Gaussian time series model over percent completion in time. Cross validation is computed with respect to a single participant over a model trained over all other participants, then computed as the mean when this procedure is repeated for all participants. The shaded region represents one standard deviation from the mean cross-validated accuracy.}
    \label{fig:pred_performance}
    \vspace{-10pt}
\end{figure*}

The former set of experiments have demonstrated that a robot using a gaze-based head-turn cue can signal its intention to a person. The most recent experiment in our laboratory on the topic of social navigation is an attempt to make inroads on the inverse of that task -— having the robot react to a person’s gaze in order to get out of the way -— by making predictions of human walking motions based on gaze.

Holman et al \cite{holman2021watch} present a study, inspired by the experiment in Unhelkar et al. \cite{unhelkar2015human}, in which participants wear a virtual reality headset with an embedded eye tracker and walk through a simulated room towards a goal that they are instructed to reach. For each trial, participants are first instructed to walk along a straight path towards position “A”, a target $1$m directly in front of their starting position. Upon reaching position “A,” participants proceed to one of five goals placed $4$m in front of the participant’s starting position, labeled $1–5$, and placed $1$m horizontally apart from each other. The purpose of navigating to position “A” before Goals $1–5$ is to avoid conflating the effects of beginning to walk with the measured effects of the study.
 
A total of $7$ participants ($6$ male, $1$ female) ranging in age from $19–31$ (mean $22.7$) participated in this study. Study participation was limited to researchers working in the laboratory, as this study was conducted during the COVID-19 pandemic.

A multivariate Gaussian time series prediction algorithm was trained on subsets of the data collected during the pedestrian's journey, and is used to predict the final goal of their path. To extract meaningful results from the small amount of data collected, the accuracy of this model is tested via cross-validation. Results can be seen in Figure \ref{fig:pred_performance}. The top line in  Figure \ref{fig:pred_performance} indicates the performance of gaze yaw plus the position of the participant in indicating the participant's final navigational goal, showing that this model predicts their motion goal earlier than all other tested cues. While this study is limited, in the fact that it only predicts motion toward a discrete goal, it represents an inroad towards reading gaze for social navigation.



\section{Conclusion} 
\label{sec:conclusion}
This paper presents some of our efforts towards the design of a robot that can navigate in a social context while conveying its intention using a gaze-based social cue and reading the gaze of nearby pedestrians. We are currently in the process of designing a system which moves these cues beyond the confines of the hallway which we constructed for these experiments and into the real world.
Here, we list a few facets of social navigation that we believe also bear further study.
\begin{enumerate}
    \item \textbf{Up-Close Deconflicting Interactions}: While most human interactions when navigating in a crowd are seamless, there are still cases where there is a conflict and the joint navigation needs to be mediated. In such cases, pedestrians make eye contact or even use verbal communication to resolve the navigational conflict \cite{nummenmaa2009ll, unhelkar2015human, hart2020hallway}. These behaviors that are either instinctive or socially learned by people, and will need to be incorporated into a social robot's behavior.
    \item \textbf{Context Understanding}: Navigation in a familiar place like one's home will results with very different gaze and motion patterns than navigation in an open mall or in a hospital. For example, an early sociological study showed that people tend to move in small groups rather than alone, but that the group size distribution highly depends on context \cite{coleman1961equilibrium}. In order to be able to leverage the gaze of pedestrians, the robot should be aware of the context of the interaction.
    \item \textbf{Cultural Differences}: Different countries have different social norms when navigating in a crowd. It was common in our human study (see Hart et al. \cite{hart2020hallway} for details) for people to shift to the right in order to avoid other pedestrians. However in other locations, people might shift to the left or to the direction that they are already more oriented towards. 
\end{enumerate}

As we continue to develop systems for social navigation, we expect to be able to handle a richer set of features handing a wider variety of situations and conforming to factors such as external context and cultural norms.

\section*{Acknowledgments}
This work has taken place in the Learning Agents Research
Group (LARG) at UT Austin.  LARG research is supported in part by NSF
(CPS-1739964, IIS-1724157, NRI-1925082), ONR (N00014-18-2243), FLI
(RFP2-000), ARO (W911NF-19-2-0333), DARPA, Lockheed Martin, GM, and
Bosch.  Peter Stone serves as the Executive Director of Sony AI
America and receives financial compensation for this work.  The terms
of this arrangement have been reviewed and approved by the University
of Texas at Austin in accordance with its policy on objectivity in
research.


\bibliographystyle{plainnat}
\bibliography{references}

\begin{thebibliography}{27}
\providecommand{\natexlab}[1]{#1}
\providecommand{\url}[1]{\texttt{#1}}
\expandafter\ifx\csname urlstyle\endcsname\relax
  \providecommand{\doi}[1]{doi: #1}\else
  \providecommand{\doi}{doi: \begingroup \urlstyle{rm}\Url}\fi

\bibitem[Baraka and Veloso(2018)]{Baraka2018}
Kim Baraka and Manuela~M. Veloso.
\newblock Mobile service robot state revealing through expressive lights:
  Formalism, design, and evaluation.
\newblock \emph{International Journal of Social Robotics}, 10\penalty0
  (1):\penalty0 65--92, Jan 2018.

\bibitem[Bera et~al.(2017)Bera, Randhavane, Prinja, and
  Manocha]{bera2017sociosense}
Aniket Bera, Tanmay Randhavane, Rohan Prinja, and Dinesh Manocha.
\newblock Sociosense: Robot navigation amongst pedestrians with social and
  psychological constraints.
\newblock In \emph{Proceedings of the IEEE/RSJ International Conference on
  Intelligent Robots and Systems (IROS)}, pages 7018--7025, Vancouver, BC,
  Canada, September 2017.

\bibitem[Charalampous et~al.(2017)Charalampous, Kostavelis, and
  Gasteratos]{charalampous2017recent}
Konstantinos Charalampous, Ioannis Kostavelis, and Antonios Gasteratos.
\newblock Recent trends in social aware robot navigation: A survey.
\newblock \emph{Robotics and Autonomous Systems}, 93:\penalty0 85--104, April
  2017.

\bibitem[Chen et~al.(2017)Chen, Everett, Liu, and How]{chen2017socially}
Yu~Fan Chen, Michael Everett, Miao Liu, and Jonathan~P How.
\newblock Socially aware motion planning with deep reinforcement learning.
\newblock In \emph{Proceedings of the IEEE/RSJ International Conference on
  Intelligent Robots and Systems (IROS)}, pages 1343--1350, Vancouver, BC,
  Canada, September 2017.

\bibitem[Coleman and James(1961)]{coleman1961equilibrium}
James~S Coleman and John James.
\newblock The equilibrium size distribution of freely-forming groups.
\newblock \emph{Sociometry}, 24\penalty0 (1):\penalty0 36--45, March 1961.

\bibitem[Fernandez et~al.(2018)Fernandez, John, Kirmani, Hart, Sinapov, and
  Stone]{fernandez2018passive}
Rolando Fernandez, Nathan John, Sean Kirmani, Justin Hart, Jivko Sinapov, and
  Peter Stone.
\newblock Passive demonstrations of light-based robot signals for improved
  human interpretability.
\newblock In \emph{Proceedings of the IEEE International Symposium on Robot and
  Human Interactive Communication (RO-MAN)}, pages 234--239, Nanjing, China,
  August 2018.

\bibitem[Gockley et~al.(2007)Gockley, Forlizzi, and
  Simmons]{gockley2007natural}
Rachel Gockley, Jodi Forlizzi, and Reid Simmons.
\newblock Natural person-following behavior for social robots.
\newblock In \emph{Proceedings of the ACM/IEEE International Conference on
  Human-robot Interaction (HRI)}, pages 17--24, Arlington, VA, USA, March 2007.

\bibitem[Hart et~al.(2020)Hart, Mirsky, Xiao, Tejeda, Mahajan, Goo, Baldauf,
  Owen, and Stone]{hart2020hallway}
Justin Hart, Reuth Mirsky, Xuesu Xiao, Stone Tejeda, Bonny Mahajan, Jamin Goo,
  Kathryn Baldauf, Sydney Owen, and Peter Stone.
\newblock Using human-inspired signals to disambiguate navigational intentions.
\newblock In \emph{Proceedings of the International Conference on Social
  Robotics (ICSR)}, pages 320--331, Golden, Colorado, USA, November 2020.

\bibitem[Helbing and Molnar(1995)]{helbing1995social}
Dirk Helbing and Peter Molnar.
\newblock Social force model for pedestrian dynamics.
\newblock \emph{Physical review E}, 51\penalty0 (5):\penalty0 4282--4286, May
  1995.

\bibitem[Henry et~al.(2010)Henry, Vollmer, Ferris, and Fox]{henry2010learning}
Peter Henry, Christian Vollmer, Brian Ferris, and Dieter Fox.
\newblock Learning to navigate through crowded environments.
\newblock In \emph{Proceedings of the IEEE International Conference on Robotics
  and Automation (ICRA)}, pages 981--986, Anchorage, Alaska, USA, May 2010.

\bibitem[Holman et~al.(2021)Holman, Anwar, Singh, Tec, Hart, and
  Stone]{holman2021watch}
Blake Holman, Abrar Anwar, Akash Singh, Mauricio Tec, Justin Hart, and Peter
  Stone.
\newblock Watch where you’re going! gaze and head orientation as predictors
  for social robot navigation.
\newblock In \emph{Proceedings of the IEEE International Conference on Robotics
  and Automation (ICRA)}, pages 6183--6190. IEEE, 2021.

\bibitem[Khambhaita et~al.(2016)Khambhaita, Rios-Martinez, and
  Alami]{khambhaita2016head}
Harmish Khambhaita, Jorge Rios-Martinez, and Rachid Alami.
\newblock Head-body motion coordination for human aware robot navigation.
\newblock In \emph{Proceedings of the International workshop on Human-Friendly
  Robotics (HFR 2016)}, page~8, Genoa, Italy, September 2016.

\bibitem[Khandelwal et~al.(2017)Khandelwal, Zhang, Sinapov, Leonetti, Thomason,
  Yang, Gori, Svetlik, Khante, Lifschitz, et~al.]{khandelwal2017bwibots}
Piyush Khandelwal, Shiqi Zhang, Jivko Sinapov, Matteo Leonetti, Jesse Thomason,
  Fangkai Yang, Ilaria Gori, Maxwell Svetlik, Priyanka Khante, Vladimir
  Lifschitz, et~al.
\newblock {BWIB}ots: A platform for bridging the gap between ai and
  human--robot interaction research.
\newblock \emph{The International Journal of Robotics Research (IJRR)},
  36\penalty0 (5-7):\penalty0 635--659, February 2017.

\bibitem[Kitagawa et~al.(2021)Kitagawa, Liu, and Kanda]{Ryo2021Omni}
Ryo Kitagawa, Yuyi Liu, and Takayuki Kanda.
\newblock Human-inspired motion planning for omni-directional social robots.
\newblock In \emph{Proceedings of the ACM/IEEE International Conference on
  Human-Robot Interaction (HRI)}, pages 34--42, Boulder, CO, USA, 2021.

\bibitem[Kruse et~al.(2013)Kruse, Pandey, Alami, and Kirsch]{kruse2013human}
Thibault Kruse, Amit~Kumar Pandey, Rachid Alami, and Alexandra Kirsch.
\newblock Human-aware robot navigation: A survey.
\newblock \emph{Robotics and Autonomous Systems}, 61\penalty0 (12):\penalty0
  1726--1743, December 2013.

\bibitem[Mirsky et~al.(2021)Mirsky, Xiao, Hart, and
  Stone]{mirsky2021prevention}
Reuth Mirsky, Xuesu Xiao, Justin~W. Hart, and Peter Stone.
\newblock Prevention and resolution of conflicts in social navigation - a
  survey.
\newblock \emph{CoRR}, abs/2106.12113, 2021.
\newblock URL \url{https://arxiv.org/abs/2106.12113}.

\bibitem[Musse and Thalmann(1997)]{musse1997model}
Soraia~Raupp Musse and Daniel Thalmann.
\newblock A model of human crowd behavior: Group inter-relationship and
  collision detection analysis.
\newblock In \emph{Computer Animation and Simulation: Proceedings of the
  Eurographics Workshop}, pages 39--51, Budapest, Hungary, September 1997.

\bibitem[Norman(2009)]{norman2009design}
Don Norman.
\newblock \emph{The design of future things}.
\newblock Basic books, 2009.

\bibitem[Nummenmaa et~al.(2009)Nummenmaa, Hy{\"o}n{\"a}, and
  Hietanen]{nummenmaa2009ll}
Lauri Nummenmaa, Jukka Hy{\"o}n{\"a}, and Jari~K Hietanen.
\newblock I'll walk this way: Eyes reveal the direction of locomotion and make
  passersby look and go the other way.
\newblock \emph{Psychological Science}, 20\penalty0 (12):\penalty0 1454--1458,
  December 2009.

\bibitem[Okal and Arras(2016)]{okal2016learning}
Billy Okal and Kai~O Arras.
\newblock Learning socially normative robot navigation behaviors with bayesian
  inverse reinforcement learning.
\newblock In \emph{Proceedings of the IEEE International Conference on Robotics
  and Automation (ICRA)}, pages 2889--2895, Stockholm, Sweden, May 2016.

\bibitem[Patla et~al.(1999)Patla, Adkin, and Ballard]{Patla1999}
A.~E. Patla, A.~Adkin, and T.~Ballard.
\newblock Online steering: coordination and control of body center of mass,
  head and body reorientation.
\newblock \emph{Experimental Brain Research}, 129\penalty0 (4):\penalty0
  629--634, Dec 1999.

\bibitem[Ratsamee et~al.(2012)Ratsamee, Mae, Ohara, Takubo, and
  Arai]{ratsamee2012modified}
Photchara Ratsamee, Yasushi Mae, Kenichi Ohara, Tomohito Takubo, and Tatsuo
  Arai.
\newblock Modified social force model with face pose for human collision
  avoidance.
\newblock In \emph{Proceedings of the ACM/IEEE international conference on
  Human-Robot Interaction (HRI)}, pages 215--216, Boston, MA, USA, March 2012.

\bibitem[Sisbot et~al.(2007)Sisbot, Marin-Urias, Alami, and
  Simeon]{sisbot2007human}
Emrah~Akin Sisbot, Luis~F Marin-Urias, Rachid Alami, and Thierry Simeon.
\newblock A human aware mobile robot motion planner.
\newblock \emph{IEEE Transactions on Robotics (T-RO)}, 23\penalty0
  (5):\penalty0 874--883, October 2007.

\bibitem[Strassner and Langer(2005)]{strassner2005virtual}
Johannes Strassner and Marion Langer.
\newblock Virtual humans with personalized perception and dynamic levels of
  knowledge.
\newblock \emph{Computer Animation and Virtual Worlds}, 16\penalty0
  (3--4):\penalty0 331--342, September 2005.

\bibitem[Takayama et~al.(2011)Takayama, Dooley, and Ju]{takayama2011expressing}
Leila Takayama, Doug Dooley, and Wendy Ju.
\newblock Expressing thought: improving robot readability with animation
  principles.
\newblock In \emph{Proceedings of the ACM/IEEE International Conference on
  Human-Robot Interaction (HRI)}, pages 69--76, Lausanne, Switzerland, March
  2011.

\bibitem[Tamura et~al.(2010)Tamura, Fukuzawa, and Asama]{tamura2010smooth}
Yusuke Tamura, Tomohiro Fukuzawa, and Hajime Asama.
\newblock Smooth collision avoidance in human-robot coexisting environment.
\newblock In \emph{Proceedings of the IEEE/RSJ International Conference on
  Intelligent Robots and Systems (IROS)}, pages 3887--3892, Taipei, Taiwan,
  October 2010.

\bibitem[Unhelkar et~al.(2015)Unhelkar, P{\'e}rez-D'Arpino, Stirling, and
  Shah]{unhelkar2015human}
Vaibhav~V Unhelkar, Claudia P{\'e}rez-D'Arpino, Leia Stirling, and Julie~A
  Shah.
\newblock Human-robot co-navigation using anticipatory indicators of human
  walking motion.
\newblock In \emph{Proceedings of the IEEE International Conference on Robotics
  and Automation (ICRA)}, pages 6183--6190, Seattle, WA, USA, May 2015.

\end{thebibliography}

\end{document}